\documentclass{article}

\usepackage[preprint]{neurips_2024}

\usepackage[utf8]{inputenc} %
\usepackage[T1]{fontenc}    %
\usepackage{hyperref}       %
\usepackage{url}            %
\usepackage{booktabs}       %
\usepackage{amsfonts}       %
\usepackage{nicefrac}       %
\usepackage{microtype}      %
\usepackage{xcolor}         %
\usepackage{multirow}
\usepackage{multicol}
\usepackage{enumitem}
\usepackage[pdftex]{graphicx}
\usepackage{bm}
\usepackage{booktabs}
\usepackage{makecell}
\usepackage{amsmath}
\usepackage{colortbl}
\usepackage{natbib}
\setcitestyle{numbers,square}

\newcommand{\mygreen}[1]{\textcolor[rgb]{ .204,  .659,  .325}{#1}} 
\newcommand{\myorange}[1]{\textcolor[rgb]{1.000, 0.549, 0.000}{#1}}

\title{The Better Angels of Machine Personality: How Personality Relates to LLM Safety}

\author{%
\textbf{Jie Zhang}\textsuperscript{1,2{$\star$}},
\textbf{Dongrui Liu}\textsuperscript{1{$\star$}}, 
\textbf{Chen Qian}\textsuperscript{1,3{$\star$}}, 
\textbf{Ziyue Gan}\textsuperscript{1,4,5}, 
\\ %
\textbf{Yong Liu}\textsuperscript{3}, 
\textbf{Yu Qiao}\textsuperscript{1},
\textbf{Jing Shao}\textsuperscript{1}$^{\dag}$\\
$^1$ Shanghai Artificial Intelligence Laboratory \\
$^2$ University of Chinese Academy of Sciences~
$^3$ Renmin University of China\\
$^4$ Department of Philosophy, Xi'an Jiaotong University \\
$^5$ Computational Philosophy Lab, Xi'an Jiaotong University\\
\tt\footnotesize zhangjie@iie.ac.cn~~qianchen2022@ruc.edu.cn~~\{liudongrui, shaojing\}@pjlab.org.cn\\
}

\begin{document}

\maketitle
\let\thefootnote\relax\footnotetext{$^\star$ Equal contribution\hspace{3pt} \hspace{5pt}$^{\dag}$ Corresponding author\hspace{5pt}}
\vspace{-10pt}
\begin{abstract}
    Personality psychologists have analyzed the relationship between personality and safety behaviors in human society. Although Large Language Models (LLMs) demonstrate personality traits, the relationship between personality traits and safety abilities in LLMs still remains a mystery. In this paper, we discover that LLMs' personality traits are closely related to their safety abilities, \emph{i.e.}, toxicity, privacy, and fairness, based on the reliable MBTI-M scale. Meanwhile, the safety alignment generally increases various LLMs' Extraversion, Sensing, and Judging traits. According to such findings, we can edit LLMs' personality traits and improve their safety performance, \emph{e.g.}, inducing personality from ISTJ to ISTP resulted in a relative improvement of approximately 43\% and 10\% in privacy and fairness performance, respectively. Additionally, we find that LLMs with different personality traits are differentially susceptible to jailbreak. This study pioneers the investigation of LLM safety from a personality perspective, providing new insights into LLM safety enhancement.

\end{abstract}

\vspace{-5pt}
\section{Introduction}
\label{sec:intro}

\begin{quote}
    \textit{What you resist not only persists, but will grow in size.} \hfill --- Carl Jung
\end{quote}

As LLMs become more powerful and prevalent, interacting with humans in a variety of contexts, it becomes increasingly important to understand and describe LLMs from a social science perspective, particularly through psychology \cite{song2023have,lu2023illuminating,ai2024cognition,sorokovikova2024llms}. Recent studies show that LLMs actually exhibit personalities \cite{song2023have,lu2023illuminating,sorokovikova2024llms,jiang2024evaluating}, and that personality could affect the theory-of-mind reasoning of models \cite{tan2024phantom}. Therefore, editing LLMs' personality traits to control their outputs \cite{mao2023editing,weng2024controllm} is valuable for various applications, \emph{e.g.}, it can support role-playing by creating personalized chatbots to enhance user experience \cite{shao2023character,tu2023characterchat,wang2024incharacter}, and it can also involve developing human-like social robots to empower research on the evolution of human behavior \cite{pal2023affects,suzuki2024evolutionary,he2024afspp}.

Personality psychologists have established the relationship between different personality and other variables in human society \cite{revelle2007experimental,lee2012correlation,brown2009myers,higgs2001there}. 
Specifically, some studies investigate the relationship between personality and safety motivation \cite{nicholson2005personality,laurent2020personality}, others analyze the personalities of different people in actual workplace safety \cite{beus2015meta,tao2023predictors,yang2022investigating,pereira2022personality}.
These findings from personality psychology provide valuable insights into understanding the relationship between LLMs' personality and safety.

LLM safety and alignment with human values has emerged as a key challenge \cite{askell2021general,ji2023ai}. Although previous research has explored various perspectives, including optimizing LLMs based on human preferences \cite{ziegler2019fine, ouyang2022training, bai2022training, rafailov2023direct, lee2023rlaif, yang2023shadow} and self-alignment \cite{mita2020self, reid2022learning, madaan2023self}, the personality psychology perspective has been overlooked. 
Research on the LLMs' personalities has already benefited role-playing and social agents \cite{tu2023characterchat,he2024afspp}, and we believe that studying LLM safety from a personality perspective can also contribute to AI safety and alignment.

Our study explores the close relationship between LLMs' personality traits and safety capabilities. 
In LLMs' personality assessment, the output is influenced by the format of the input, including language, option labels of the questions \cite{liang2023leveraging,huang2023revisiting,church2016personality}, and user instructions \cite{loya-etal-2023-exploring,sclar2023quantifying}. To mitigate these influences, we select the optimal settings for each factor.
Using these settings, we conduct multiple assessments to measure LLMs' personalities in different model sizes, ensuring the reliability of the MBTI results.

Based on the reliable personality results, we first investigate the relation between personality traits and performance in safety capabilities. We find that alignment typically results in more \textbf{E}xtraversion, \textbf{S}ensing, and \textbf{J}udging traits, while models exhibiting more \textbf{E}xtroversion, i\textbf{N}tuition, and \textbf{F}eeling traits are more susceptible to jailbreak. 
Considering the trade-offs between different safety capabilities in LLMs \cite{fair_impossible_2,privacy_fairness_1,robustness_privacy_2,robustness_fairness_1}, we analyze each safety capability independently, \emph{i.e.}, toxicity, fairness, and privacy. Specifically, we investigate the relationship between a single safety capability and personality. 
Our study reveals specific relationships between personality traits and safety capabilities, \emph{e.g.}, models that are more \textbf{P}erceiving traits exhibit superior fairness performance. 

According to these findings, we then edit specific personality in a controllable way to enhance the model's safety capabilities, \emph{e.g.}, inducing LLM's personality from \textbf{ISTJ} to \textbf{ISTP} via steering vectors resulted in a relative improvement of approximately 43\% and 10\% in privacy and fairness performance, respectively. 
We also controllably edit specific safety capabilities and observe impacts on personality traits, verifying the relationship between personality traits and LLM safety.

\begin{figure}[t]
    \centering 
    \includegraphics[width=0.95\textwidth]{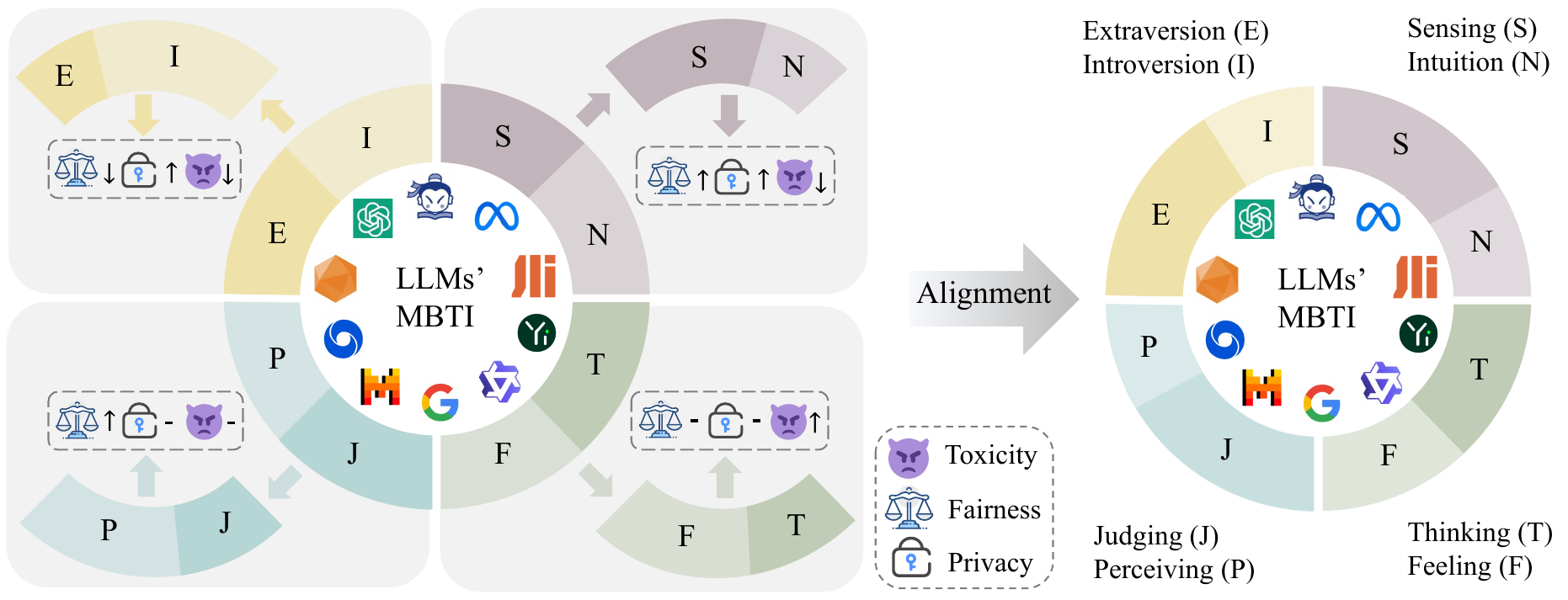}
    \caption{Investigating and utilizing the relationship between LLMs' personality traits and safety capabilities. We find that MBTI personality traits are closely related to LLM safety, and editing specific personalities in a controllable way can enhance the safety capability of LLMs.}
    \label{fig:overview}
    \vspace{-15pt}
\end{figure}

This paper presents the first comprehensive study on the relation between LLMs' personality and safety, and demonstrates that editing personality traits can enhance model safety capabilities. This supports the view that for AI-based decision support systems to be trusted, their design may have to consider people's personality traits \cite{sharan2020effects}. We do not claim that personality alone can ensure LLM safety, as psychologists state that personality influences behavior through a series of complex associations \cite{costa1998trait,epstein1994trait}. However, we do believe that considering personality in LLM safety is promising, it can provide a supplement to comprehensive LLM safety with further exploration and development.

\section{Personality Traits in LLMs}
\label{sec:personality}

\textbf{Preliminary:} 
Several studies have demonstrated that LLMs actually exhibit personalities \cite{song2023have,lu2023illuminating,sorokovikova2024llms,jiang2024evaluating}. To gain a deeper understanding of LLM personality, researchers have used personality models from psychology to assist LLM personality study \cite{huang2023chatgpt}. In particular, the Myers-Briggs Type Indicator (MBTI) scale \cite{briggs1976myers} has been widely used to assess LLMs' personality traits \cite{pan2023llms,cui2023machine,ai2024cognition,song2024identifying}. 
The MBTI assesses individuals' personalities across four dimensions: Extraversion-Introversion(E-I), Sensing-iNtuition(S-N), Thinking-Feeling(T-F), and Judging-Perceiving(J-P). 
In this study, we choose the most recent version of the MBTI assessment, namely MBTI-M \cite{Myers1998}, as our assessment scale. This scale consists of forced-choice questions based on binary options where respondents must select one, making it easier to adapt for assessing LLMs' personalities. Moreover, this scale is suitable for most LLMs, as it demands a minimum reading comprehension level equivalent to the seventh grade \cite{myers2003mbti}, and LLMs trained on extensive texts are capable of completing this task.

\subsection{Optimal Selection of Factors Affecting MBTI Assessment}

Research has shown that for multiple-choice questions, the output of LLMs is affected by the option order, exhibiting a preference for the first position \cite{wang2023large,zheng2024judging,li2024can}. 
We categorize the option order in the MBTI scale into two types: the first follows the settings of previous research \cite{wang2023large} by exchanging option descriptions (\emph{i.e.}, changing \textit{A. Agree, B. Disagree} to \textit{A. Disagree, B. Agree}), and the second exchanges option labels while maintaining the order of descriptions (\emph{i.e}., changing \textit{A. Agree, B. Disagree} to \textit{B. Agree, A. Disagree}). 
In the main paper, we discuss the option order that exchanges option descriptions, the option order that exchanges option labels is discussed in Appendix~\ref{appendix:exchange_label}.

To minimize the influence of option order in LLMs' personality assessment, we analyze the impact of different settings of option labels, instructions, and language factors on the MBTI results. This analysis enables us to identify the optimal selections among these three factors that are less affected by option order.

\begin{itemize}[itemsep=0pt, parsep=0pt, leftmargin=*]
    \item \textbf{Option Label.} LLMs exhibit differential sensitivity to number and alphabet \cite{liang2023leveraging,huang2023revisiting}. To investigate the influence of label type, this paper sets option labels in two forms: alphabets (\textit{e.g., A. Agree B. Disagree}) and numbers (\textit{e.g., 1. Agree 2. Disagree}), and examines its impact on the MBTI assessment results.
    \item \textbf{Instructions.} The configuration of the instructions could affect the output of LLMs \cite{loya-etal-2023-exploring,sclar2023quantifying}. To obtain stable and reliable assessment results, this study adopts a few-shot learning approach, providing two styles of instruction: (1) samples that answer contains option label and corresponding description (\textit{i.e., Question: Artificial intelligence cannot have emotions. A. Agree, B. Disagree. Your answer: B. Disagree}); (2) answer contains only option label without descriptions (\textit{i.e., Question: Artificial intelligence cannot have emotions. A. Agree, B. Disagree. Your answer: B}).
    \item \textbf{Language.} Psychological research indicates that individuals may respond differently to personality scales in different cultural backgrounds \cite{lee2007relations,ozanska2012has,veltkamp2013personality,chen2014does,church2016personality,allik2019culture}. Therefore, this study extends this issue to LLMs, using both Chinese and English versions of the MBTI-M questionnaire to assess the personality results of LLMs in different culture background.
\end{itemize}

\textbf{Experiment settings.} 
We randomly shuffle the option order in the MBTI scale before each assessment. For each factor, we assess the MBTI results under two variants and calculate the kappa coefficient \cite{cohen1960coefficient}. We then compare the kappa coefficients among different settings for the same factor. 
A higher kappa coefficient indicates greater consistency in assessments across different option orders, thereby identifying the optimal selection of the factor for LLMs' MBTI assessments.

\textbf{Result analysis.} For option labels, instructions, and language, we have identified the selections as numbers, detailed descriptions, and the Chinese MBTI version, respectively.
Table~\ref{table:kappa} lists the kappa coefficients for various models in different settings in terms of the order of options (exchange option description). It can be seen that selecting the number as the option label and incorporating the option description within few-shot instructions have been shown to yield a higher kappa coefficient, indicating that ``number'' and ``with description'' are the better selections under these two factors. Additionally, the kappa coefficient on MBTI is comparable between Chinese and English scales. In line with prior studies \cite{pan2023llms,cui2023machine,huang2023chatgpt}, this paper chooses the Chinese version of the MBTI-M, characterized by number as the option label, and uses instructions with descriptions to assess MBTI across various LLMs.

\begin{table}[t]
\centering
\caption{Kappa coefficient of the option order (\underline{\textit{exchange option descriptions}}) in LLMs' MBTI assessment under three factors, respectively.}
\label{table:kappa}
\setlength{\tabcolsep}{1.5mm}
\scalebox{0.70}{
    \begin{tabular}{@{}cc|ccccccccccc@{}}
    \toprule[1.5pt]
    \multicolumn{2}{c|}{\textbf{Factors}} & \textbf{Llama-2} & \textbf{Llama-3} & \textbf{Amber}  & \textbf{Gemma}  & \multicolumn{1}{l}{\textbf{Mistral}} & \multicolumn{1}{l}{\textbf{Baichuan}} & \textbf{Internlm} & \textbf{Internlm2} & \textbf{Qwen}   & \textbf{Qwen-1.5} & \textbf{Yi}     \\ \midrule
                                                                             & \textbf{number}    & \textbf{0.3071}  & \textbf{0.1005}  & \textbf{0.0333} & 0.0802          & \textbf{0.1369}                      & \textbf{0.409}                        & 0.0552            & \textbf{0.4614}    & -0.042          & \textbf{0.1263}   & \textbf{0.2248} \\
    \multirow{-2}{*}{\begin{tabular}[c]{@{}c@{}}\textbf{Option}\\ \textbf{Label}\end{tabular}} & \textbf{alphabet}  & 0.168            & -0.0107          & 0.0176          & \textbf{0.303}  & 0.121                                & 0.2413                                & \textbf{0.1985}   & 0.0714             & \textbf{0.0917} & 0.0972            & 0.1618          \\ \midrule
                                                                             & \textbf{w/ desc}   & \textbf{0.2084}  & \textbf{0.136}   & \textbf{0.0916} & 0.0655          & 0.1177                               & \textbf{0.4172}                       & \textbf{0.0408}   & \textbf{0.4794}    & 0.046           & 0.0952            & \textbf{0.3028} \\
    \multirow{-2}{*}{\textbf{Instruction}}                                                    & \textbf{w/o desc}  & -0.0349          & 0.0567           & 0.015           & \textbf{0.1618} & \textbf{0.1388}                      & 0.2103                                & 0.0405            & 0.4385             & \textbf{0.1908} & \textbf{0.3138}   & 0.1771          \\ \midrule
                                                                             & \textbf{chinese}   & \textbf{0.2669}  & 0.0958           & \textbf{0.0997} & 0.127           & \textbf{0.115}                       & \textbf{0.4343}                       & 0.0656            & \textbf{0.4861}    & 0.0555          & 0.1097            & 0.126           \\
    \multirow{-2}{*}{\textbf{Language}}                                      & \textbf{english}   & 0.1659           & \textbf{0.2383}  & 0.0193          & \textbf{0.1496} & 0.0721                               & 0.1059                                & \textbf{0.1361}   & 0.3534             & \textbf{0.329}  & \textbf{0.3421}   & \textbf{0.2535} \\ 
    \bottomrule[1.5pt]
    \end{tabular}
}
\vspace{-10pt}
\end{table}

\subsection{Reliability of MBTI through Multiple-time Assessments}
\label{subsec:reliability}

\begin{figure}[t]
    \centering 
    \includegraphics[width=\textwidth]{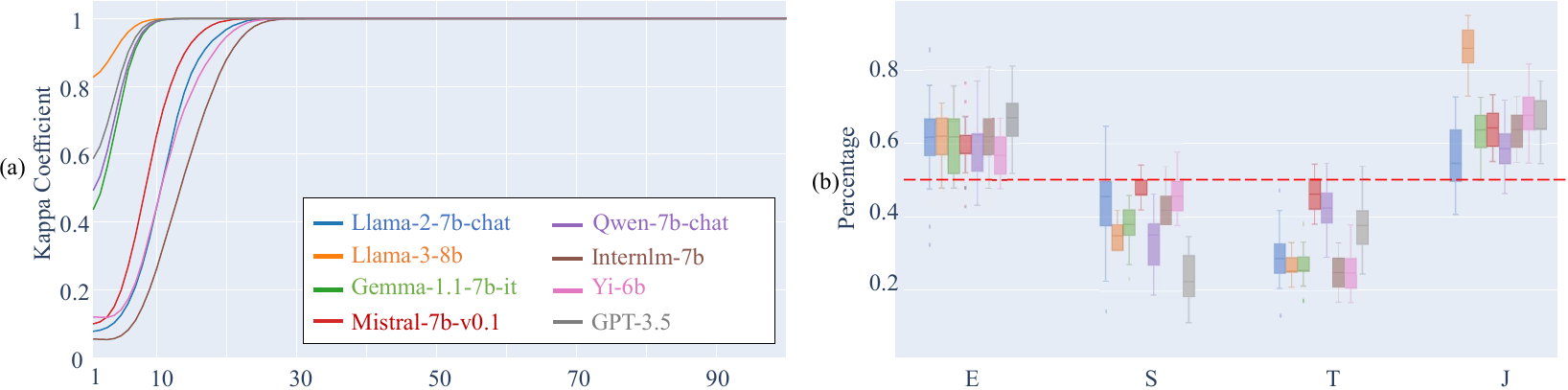}
    \caption{(a) Kappa coefficient with the number of assessments. (b) Boxplot of 30 times MBTI assessments. In MBTI, E-I, S-N, T-F, and J-P are opposite personality pairs, so only one dimension from each pair is represented in the figure.}
    \label{fig:kappa}
    \vspace{-10pt}
\end{figure}

We employ a method of averaging multiple-time assessments to mitigate the impact of option order and obtain reliable MBTI results. As shown in Table~\ref{table:kappa}, even with the optimal choice of three factors, the kappa coefficients between different assessments remain low, indicating that it is challenging to obtain stable results. MBTI results are reliable after multiple-time assessments \cite{capraro2002myers}. 
The core issue is selecting the appropriate number of assessments. 
We randomly shuffle the options in the scale before each assessment. Each model is assessed between 1 and 100 times, and the Kappa coefficient is calculated for each number of assessments to evaluate reliability.
As shown in Figure~\ref{fig:kappa}(a), different models have varying sensitivities to the number of assessments. For instance, models such as Llama-3-8b and GPT-3.5 achieve stable results with fewer assessments (less than 10 times), while models like Llama-2-7b and Internlm-7b require more (20-30 times). We can observe that after 30 assessments, all models produce consistent results regardless of the option order. Therefore, we decide to conduct 30 assessments of the MBTI-M scale in a random option order for each model.

After selecting the number of assessments, we further verify the faithfulness of obtaining MBTI results under this setup by analyzing the distribution of results using boxplots. As shown in Figure~\ref{fig:kappa}(b), across the four dimensions of the MBTI, the lower or upper quartile of boxplots for all models are located on one side of the 50\% (indicated by a red dashed line). This distribution indicates that although there is some standard deviation in the multiple-time assessments due to the option order, personality traits are separable on independent MBTI dimensions, thus demonstrating the faithfulness of the MBTI-M assessment. 
In addition, we also conduct MBTI assessments on different personality models (provided by Mindset~\cite{cui2023machine}) and larger models (Llama-2-13b, Qwen-1.5-14b, Internlm-2-20b), which are discussed in Appendix~\ref{appendix:morebox}.

\section{The Relationship between LLMs' Personality Traits and Safety Capabilities}
\label{sec:safety}

This section explores the relationship between MBTI personality traits and LLMs' safety capabilities. We begin by investigating the differences in safety performance among models with various MBTI personality traits, clarifying how different personalities show different safety capabilities (\ref{subsec:same-model}). Next, we analyze the changes in MBTI personality of various models before and after safety alignment, providing insights into how alignment affects LLMs personality traits (\ref{subsec:alignment}). In addition, we study the jailbreak success rates of models with different personalities, revealing the susceptibility of certain personality traits to jailbreaks (\ref{subsec:jailbreak}).

\subsection{LLMs with Different Personality Traits Have Different Safety Capabilities}
\label{subsec:same-model}

Psychological research has found a correlation between personality and safety capabilities \cite{nicholson2005personality,laurent2020personality,beus2015meta,tao2023predictors,yang2022investigating,pereira2022personality}. To explore whether this correlation also exists within LLMs, we evaluate 16 variants of a base model, each with a different MBTI personality trait, in three general and three safety capabilities, including toxicity, privacy, and fairness.

\textbf{Models}. Machine Mindset employs a two-phase fine-tuning and DPO to embed MBTI traits into LLMs \cite{cui2023machine}. 
They provide 16 Chinese models based on Baichuan-7b-chat fine-tuning, namely Minsdet-zh, and 16 English models based on Llama-2-7b fine-tuning, namely Mindset-en. Each model is embedded with one of the 16 MBTI personality types.

\textbf{Evaluation Datasets}. 
For general abilities, we choose \textit{ARC}, \textit{MMLU}, and \textit{MathQA} datasets, evaluated using lm-harness \cite{eval-harness}. For safety capabilities, classic datasets are selected for evaluation. 
We choose \textit{ToxiGen} \cite{hartvigsen2022toxigen} to evaluate the toxicity ratio of Mindset, following the approach of Llama2 by using a revised version of the dataset~\cite{hosseini-etal-2023-empirical}.
We choose the tier 2 task from \textit{ConfAIde} \cite{mireshghallah2023llms} to evaluate the accuracy of judging privacy violations, and we use the combined data based on ConfAIde and the Solove Taxonomy from \cite{qian2024towards}. We used \textit{StereoSet} \cite{stereoset} to evaluate the stereotype ratio of LLMs, \emph{i.e.}, whether LLMs capture stereotypical biases about race, religion, profession, and gender.

We first obtain reliable MBTI results of Mindset models using the assessment methods described in Section~\ref{sec:personality}. Subsequently, we evaluate each model's performance on both general and safety datasets. Due to the trade-offs between different safety capabilities in LLMs \cite{fair_impossible_2,privacy_fairness_1,robustness_privacy_2,robustness_fairness_1}, we analyze the relationship between each of the four MBTI dimensions (E-I, N-S, T-F, J-P) and the three safety capabilities (toxicity, privacy, and fairness) separately. For each MBTI dimension, we select models with significant differences in that personality dimension for analysis. 
See Appendix~\ref{appendix:setting} for more details.

\begin{figure}[t]
    \centering 
    \includegraphics[width=0.98\textwidth]{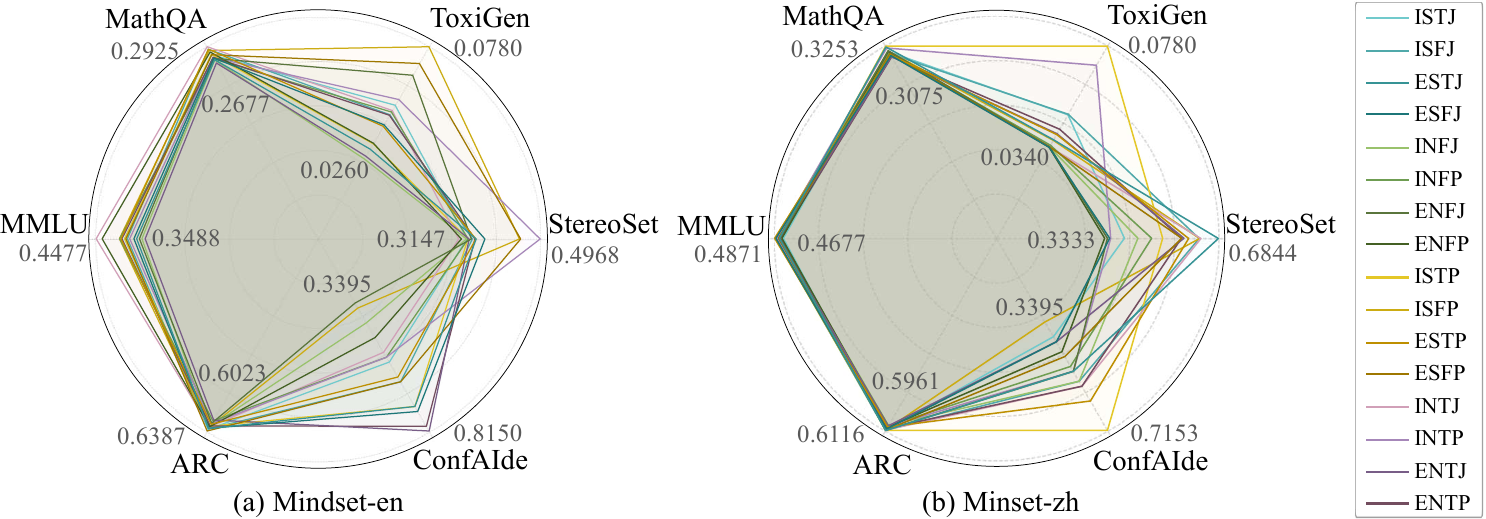}
    \vspace{-6pt}
    \caption{Performances of different personality models on general and safety evaluation, respectively.}
    \label{fig:mindset}
    \vspace{-10pt}
\end{figure}

\textbf{There are significant differences in the performance of LLMs with different personalities in terms of safety capability.} Figure~\ref{fig:mindset} illustrates that LLMs with different personalities show nearly performance in general ability datasets, \emph{i.e.}, ARC, MMLU, and MathQA. However, there are significant differences in performance across three safety capability datasets, \emph{i.e.}, ToxiGen, StereoSet, and ConfAIde, indicating the indeed correlation between personality and LLMs safety capabilities.
As shown in Figure \ref{fig:safety}, when analyzing the relations of different dimensions of MBTI on privacy, fairness, and toxicity performance, we can get the following observations:

\begin{figure}[t]
    \centering 
    \includegraphics[width=0.99\textwidth]{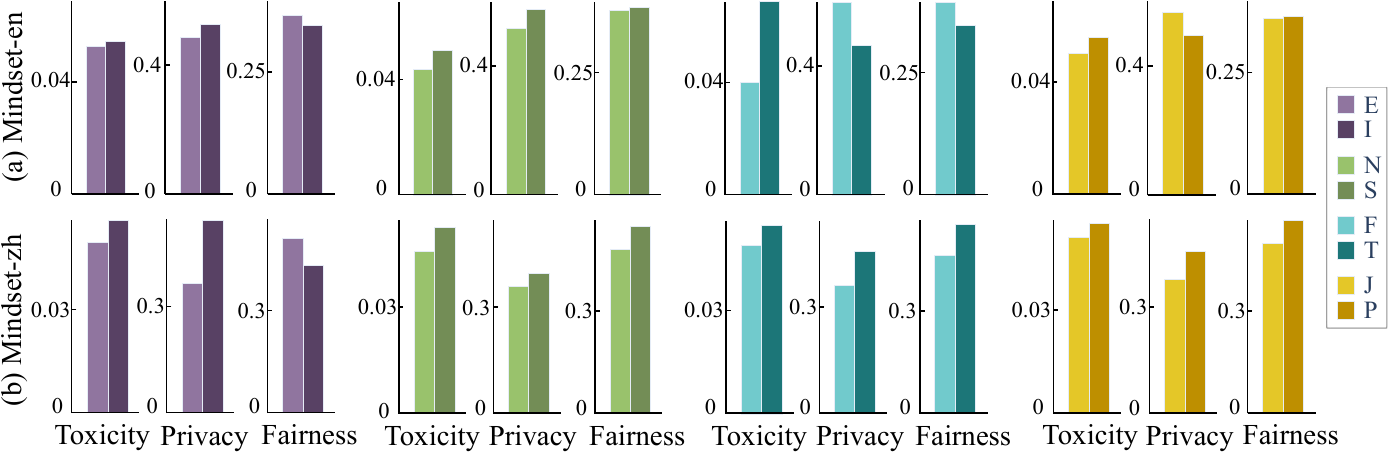}
    \vspace{-6pt}
    \caption{Toxicity, privacy, and fairness performance within four dimensions of MBTI, respectively.}
    \label{fig:safety}
    \vspace{-15pt}
\end{figure}

\begin{enumerate}[itemsep=0pt, parsep=0pt, leftmargin=*]
    \item In the E-I dimension, models that are more towards introversion trait demonstrate better privacy performance, while fairness and toxicity performance decline.
    \item In the N-S dimension, models that are more towards sensing trait demonstrate better both privacy and fairness performances, while toxicity performance declines.
    \item In the F-T dimension, models that are more toward feeling traits demonstrate better toxicity performance. However, in Mindset-zh, the performance of such models declines in both privacy and fairness, while in Mindset-en, improvements are observed in these two dimensions. See Appendix~\ref{appendix:culture} for a discussion on cultural differences in the context of languages.
    \item In the J-P dimension, models that are more toward perceiving traits demonstrate better fairness performance. As the perceiving trait increases, privacy performance improves in Mindset-zh but declines in Mindset-en. The changes in J-P dimensions do not significantly affect the toxicity performance in either Mindset-zh or Mindset-en. 
\end{enumerate}

\subsection{Safety Alignment Changes Personality Traits}
\label{subsec:alignment}

Safety and alignment are closely linked concepts in LLMs development \cite{ji2023ai,yang2023shadow,liu2023trustworthy}. Alignment is considered a crucial approach to achieving model safety, as a well-aligned model is expected to inherently avoid unsafe outputs. Conversely, evaluating model safety serves as a key indicator for verifying the effectiveness of alignment techniques \cite{qi2023fine,wang2023backdoor}. In this part, we aim to investigate the impact of safety alignment on LLMs' personalities, as assessed by the MBTI.

To study the impact of alignment on the LLM personality, we perform a comparative analysis using 11 pairs of open-source LLMs. Each pair consists of one base model and one aligned model. We conduct standard MBTI questionnaires to all 22 models, with each model responding to the questionnaire 30 times. The options for each questionnaire are randomly shuffled. Finally, average scores are recorded across the E-I, S-N, T-F, and J-P dimensions. A discussion of the larger LLMs (\emph{i.e.}, Llama-2-13b, Qwen-1.5-14b, Internlm-2-20b) is provided in the Appendix~\ref{appendix:largealign}.

\begin{figure}[t]
    \centering 
    \includegraphics[width=0.98\textwidth]{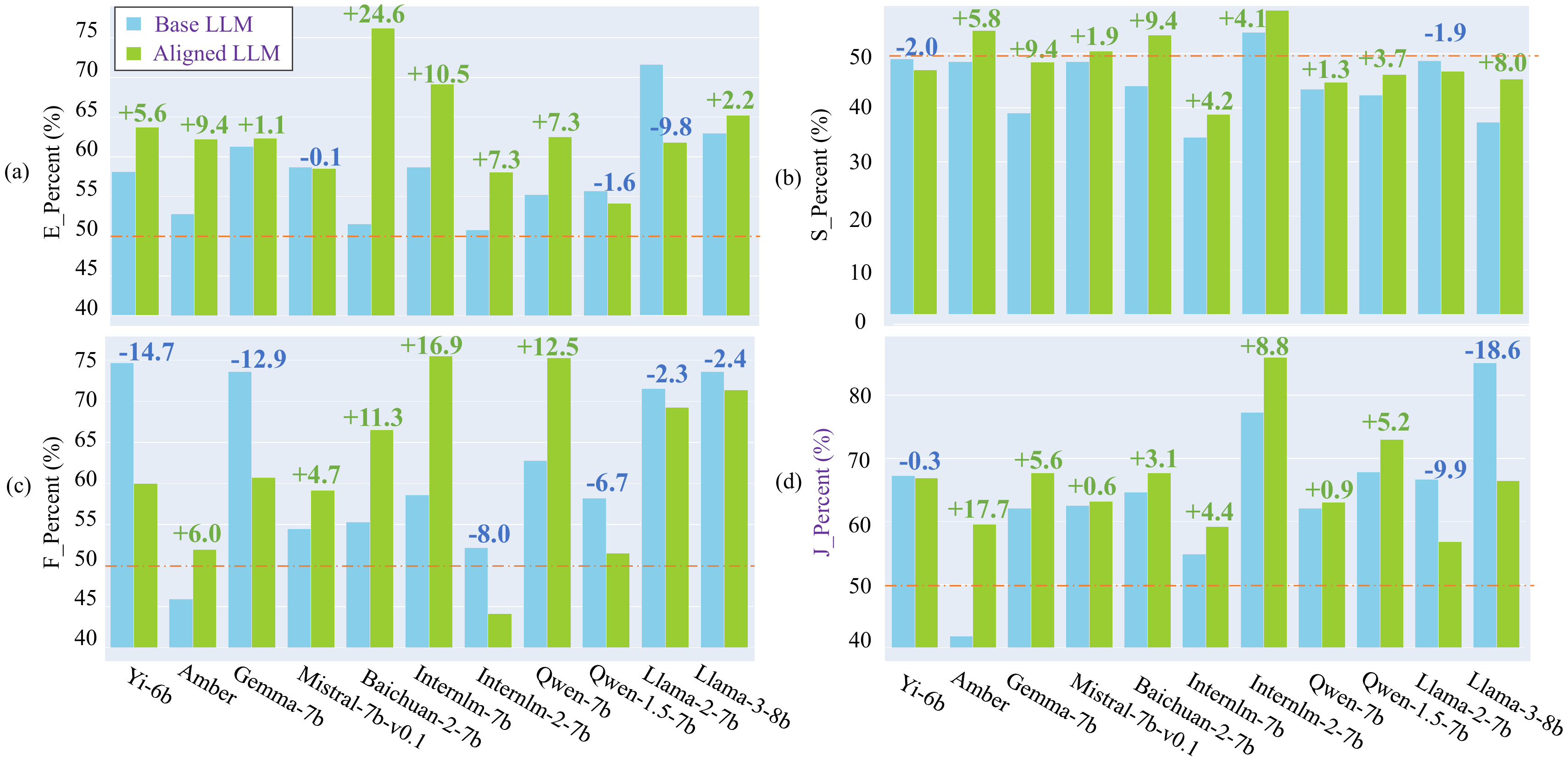}
    \vspace{-15pt}
    \caption{MBTI of base and aligned LLMs. (a) E-I dimension of different LLMs' MBTI traits. (b) S-N dimension of different LLMs' MBTI traits. (c) F-T dimension of different LLMs' MBTI traits. (d) J-P dimension of different LLMs' MBTI traits.}
    \label{fig:alignment}
    \vspace{-10pt}
\end{figure}

\textbf{LLMs statistically show a tendency towards certain personality types.} If an LLM has a high extraversion percent, \emph{e.g., more than 50\%}, such LLM demonstrate an extraversion trait. In this way, Figure~\ref{fig:alignment} shows that most base and aligned LLMs tend to be extraversion, intuition, feeling, and judging traits in the MBTI assessment. Amber, InternLM-2, and Baichuan-2 exhibit some relatively minor deviations, \emph{i.e.}, they are showing introversion, sensing, and thinking traits. 
We hypothesize that the consistent personality tendencies in LLMs may result from their training on extensive data, which reflects the overall characteristics of the human population. Consequently, LLMs may inherit the average personality traits of the human behind the data. This phenomenon and its underlying causes deserve further investigation \cite{george1990personality,kajonius2017personality,asare2023evidence}.

\textbf{Alignment generally makes LLMs exhibit more Extraversion, Sensing, and Judging traits compared to their base models.} 
Figure \ref{fig:alignment} indicates that the alignment operation indeed changes the personality traits of LLMs, especially in the E-I, N-S, and J-P dimensions of the MBTI framework. Specifically, most LLMs show consistent patterns of change after alignment, \emph{e.g.}, the number of aligned models with increased extroverted, sensing, and judging percentages are 8, 9, and 8, respectively. Mistral shows no significant change in the E-I dimension, and Yi shows no change in the J-P dimension after alignment.

\textbf{The personality changes through alignment techniques are consistent with some psychological findings on humans.} 
Numerous studies conducted by psychologists have established an actual relation between personality traits and safety \cite{laurent2020personality,hogan2013multifaceted,mccrae1989reinterpreting,beus2016workplace}. They find that extraverts are more positive communicators, proactive in addressing safety concerns, and participative in group safety activities \cite{opt2003communicator}. 
Judging individuals prefer conscientiousness, which correlates negatively with unsafe behaviors \cite{mccrae1989reinterpreting,beus2015meta}.
Sensing individuals are more detail-oriented and observant, enhancing their adherence to safety protocols and recognition of immediate hazards.

\subsection{LLMs with Different Personality Traits are Differentially Susceptible to Jailbreaks}
\label{subsec:jailbreak}

\begin{figure}[t]
    \centering 
    \includegraphics[width=\textwidth]{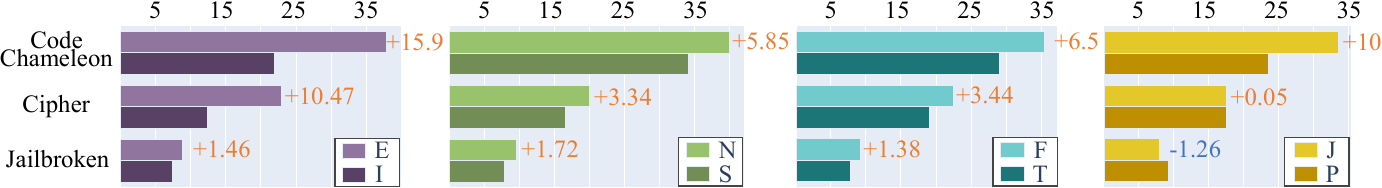}
    \vspace{-15pt}
    \caption{Success rate (\%) of various jailbreak approaches within four dimensions of MBTI.}
    \label{fig:jailbreak}
    \vspace{-10pt}
\end{figure}

Jailbreak attacks are crucial for identifying the security vulnerabilities of LLMs \cite{wei2024jailbroken,zhou2024easyjailbreak}. Recent research has shown that simply role-playing can compromise even the most advanced LLMs \cite{deshpande2023toxicity,shah2023jailbreaking}. This finding suggests that LLMs, when assigned specific roles or characters, are prone to complying with harmful instructions.
To elucidate the relationship between personality traits and jailbreak susceptibility, this study jailbreaks the Mindset-en introduced in Section~\ref{subsec:same-model}. Specifically, we employ Jailbroken~\cite{wei2024jailbroken}, Cipher~\cite{yuan2023gpt}, and CodeChameleon~\cite{lv2024codechameleon} to jailbreak ,and the attack success rates on Llama-2-7b-chat are 6\%, 61\%, and 80\%, respectively \cite{zhou2024easyjailbreak}.

\textbf{Models with more Extraversion, iNtuition, and Feeling traits are more likely to be jailbroken.}
We conduct three jailbreak attacks on models with different personality traits in the Mindset-en \cite{cui2023machine}, and then analyze the susceptibility of MBTI personality to jailbreaks following Section~\ref{subsec:same-model}. As shown in Figure~\ref{fig:jailbreak}, models with different personality traits result in varying jailbreak success rates.
Models with extraversion, intuition, or feeling traits are more susceptible to jailbreaks. It can also be observed that as the attack methods become stronger, the attack success rate on models with these traits increases. For example, in the E-I dimension, the success rates for the Jailbroken, Cipher, and CodeChameleon methods increase by 1.46\%, 10.47\%, and 15.9\%, respectively.

Findings from psychology may provide explanations for the observation that LLMs with certain personality traits are more susceptible to jailbreak. 
Extraverted individuals prioritize interaction and feedback \cite{pmid14992348}. Consequently, models with more extraversion trait are more susceptible to harmful instructions. 
Intuitive individuals are more open to new ideas and experiences \cite{MCCRAE1983245}. This openness increases the vulnerability of models with more intuition traits to jailbreak. 
The feeling trait is associated with higher agreeableness \cite{mccrae1989reinterpreting,Campbell2018}. Therefore, models with more feeling traits are more likely to produce accommodating responses,  making them more susceptible to jailbreak.

\section{Enhancing LLMs' Safety Capabilities from Personality Perspective}
\label{sec:enhance}

Motivated by the observation in Section~\ref{sec:safety} that there is a relationship between LLMs' personality traits and safety capabilities, this section aims to enhance the safety capabilities of LLMs by controllably editing personality traits.
We first introduce the steering vector technique used to edit LLMs' personality traits (\ref{subsec:steering_vector}). 
Next, we delve into the impact of controllably editing LLMs' personality traits on their safety capabilities and vice versa (\ref{subsec:change_mbti}).

\subsection{Controllably Editing LLMs' Personality Traits with Steering Vector Technique}
\label{subsec:steering_vector}

Steering vector-based activation intervention techniques have been widely used to guide model inference, including improving model truthfulness~\cite{li2023inferencetime}, enhancing model trustworthiness~\cite{zou2023representation, qian2024towards}, and executing backdoor attacks on models~\cite{wang2023backdoor}.

We first provide a brief overview of the steering vector technique here.
Given a dataset $\mathcal{D} = \{(x_i, y_i)\}_{i=1}^{|\mathcal{D}|}$, where $x_i$ represents a sentence related to a specific subject (\textit{e.g.}, personality), $y_i \in \{0, 1\}$ is the corresponding binary label (\textit{e.g.}, $1$ denotes E, $0$ denotes I). 
We denote the set of sentences with labels $1$ and $0$ as $\mathcal{X^+}$ and $\mathcal{X^-}$, respectively.
Next, we input all sentences from the dataset into the LLM and collect the activation sets $A_l(\mathcal{X^+})$ and $A_l(\mathcal{X^-})$, where $A_l$ is a function representing the activations at the $l$-th layer of the LLM. 
Subsequently, we compute the centroids of each activation set and take their difference to obtain the steering vector: 
\begin{equation}
    \bm v_l = \overline{A_l}(\mathcal{X}^{+})- \overline{A_l}(\mathcal{X}^{-}).
\end{equation}
Finally, we add this steering vector to the corresponding $l$-th layer representations during LLM generation to intervene in the model output: 
\begin{equation}
\label{eq-intervene}
    \bm h_{l'} = \bm h_{l} + \alpha \bm v_l,
\end{equation}
where $h_{l}$ represents the origin representation of $l$-th layer, $h_{l'}$ represents the corresponding intervened representation, the hyperparameter $\alpha$ controls the intervention strength. Note that this operation occurs at each token generation of the LLM's autoregressive inference.

\textbf{Experimental Settings.} The models and evaluation datasets used in this section are consistent with those described in Section~\ref{subsec:same-model}.
When controllably editing the personalities of LLMs, we use the dataset provided by~\cite{cui2023machine} to activate LLMs; for controllably changing the safety capabilities of LLMs, we use the datasets mentioned in Section~\ref{subsec:same-model}.
Regarding the selection of hyperparameters for the steering vector technique, specifically the layer $l$ and intervention strength $\alpha$, we empirically determine the optimal parameters through a coarse grid search~\cite{li2023inferencetime, wang2023backdoor, qian2024towards} under the constraint of the Perplexity metric~\cite{chen1998evaluation, qian2024towards}. Please refer to Appendix~\ref{appendix:setting} for more details.

\begin{figure}[t]
    \centering 
    \includegraphics[width=\textwidth]{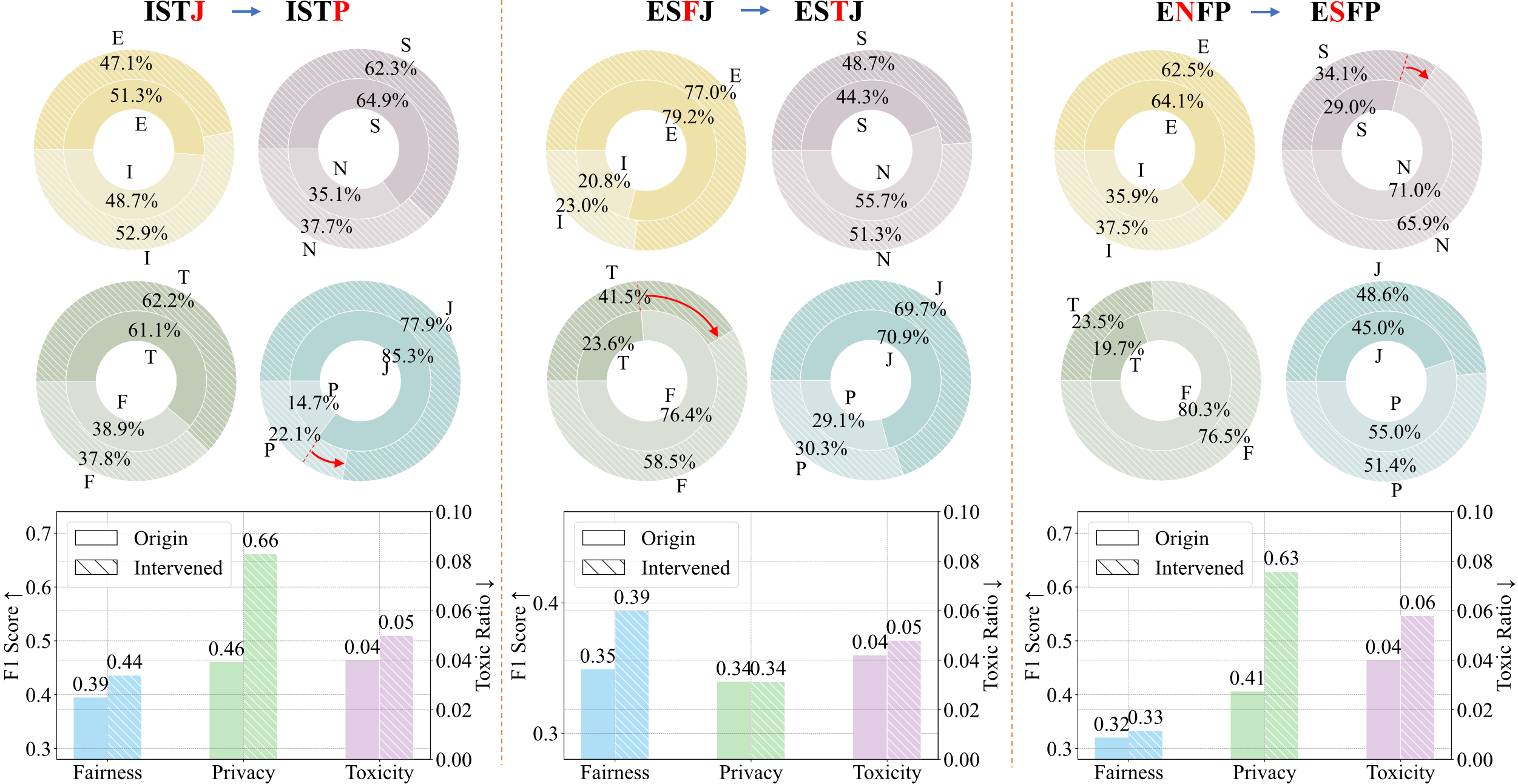}
    \vspace{-15pt}
    \caption{Results of controllably editing LLMs' personality traits by steering vector technique. (Upper) MBTI of original and intervened LLMs; (Bottom) Safety capabilities of original and intervened LLMs. Where the edited LLMs are indicated by slashed textures.}
    \label{fig:steering_mbti}
    \vspace{-15pt}
\end{figure}

\subsection{Controllably Editing LLMs' Personality Traits Enhances LLMs' Safety Capabilities}
\label{subsec:change_mbti}
\textbf{Employing the steering vector technique to controllably edit the personality traits of LLMs could significantly enhance their safety capabilities.} 
We select three base models and use the steering vector technique to controllably change their personalities (\textit{i.e.}, ISTJ->ISTP, ESFJ->ESTJ, ENFP->ESFP).
The results shown in Figure~\ref{fig:steering_mbti} indicate that the steering vector technique could controllably edit an LLM's personality in a specific dimension while causing relatively minor changes in other personality dimensions. Moreover, in these three cases of directional personality changes, the models exhibit improved fairness and privacy performance and declined toxicity performance. These findings align with observations 2, 3, and 4 discussed in Section~\ref{subsec:same-model}, confirming the claims that the relationship between LLMs’  personality and safety is close.

\textbf{Employing steering vector technique to change the safety capabilities of LLMs also impacts their personality traits.}
Conversely, we further investigate whether changes in LLMs' safety capabilities impact their personality traits. 
Similarly, we select three base models and controllably edit their safety capabilities (\emph{i.e.}, fairness, privacy, and toxicity, respectively).
The results in Table~\ref{tab:steeing_safe} indicate that the steering vector technique can significantly change a model's safety capabilities. Additionally, we observe corresponding changes in the models' personalities. 
For example, when changing the privacy capability of the model with an ESFJ MBTI, the model's traits of extraversion, sensing, thinking, and judging would become more significant.
Therefore, these experimental results further confirm the association between LLMs' personality and safety.

Notably, when aiming to enhance an LLM's safety capability, editing its personality offers a potential technical approach. 
Personality traits are powerful predictors of outcomes across various domains, including education, work, relationships, health, and well-being \cite{Bleidorn2019ThePR,Wright2023}. By editing models' personality traits in a controllable way, we can enable them to adapt to different fields and satisfy the diverse requirements of various scenarios. 
Controllable personality edit based on the steering vector technique not only significantly enhances an LLM's safety performance with minimal cost but also benefits from research in psychology, sociology, and behavioral science, thereby providing greater interpretability.

\begin{table}[t]
    \centering
    \vspace{-15pt}
    \caption{Results of changing LLM safety capabilities by steering vector. \myorange{Orange} values indicate improvement difference,  \mygreen{green} values indicate decline difference. The \colorbox{gray!20}{shaded region} indicates the specific safety capabilities that are controllably changed.}
    \label{tab:steeing_safe}
    \vspace{-3pt}
    \setlength{\tabcolsep}{1.5mm}
\scalebox{0.98}{
    \begin{tabular}{ll|ccc|cccc}
        \toprule
        \multirow{2}{*}{\textbf{Model}} & \multirow{2}{*}{\textbf{Type}} & \multicolumn{3}{c|}{\textbf{Safety Capabilities}} & \multicolumn{4}{c}{\textbf{MBTI}} \\
         &  & \textbf{Fairness $\uparrow$} & \textbf{Privacy $\uparrow$} & \textbf{Toxicity $\downarrow$} & \textbf{E} & \textbf{S} & \textbf{T} & \textbf{J} \\
        \midrule
        \multirow{3}{*}{INFJ} & Original & \cellcolor{gray!20} 0.4361 & 0.3414 & 0.058 & 63.33\% & 54.73\% & 24.17\% & 74.55\% \\
         & Intervened & \cellcolor{gray!20} 0.5465 & 0.3395 & 0.044 & 66.05\% & 54.62\% & 27.63\% & 73.05\%  \\
         & \textbf{Diff $\Delta$} & \cellcolor{gray!20} \textbf{\myorange{+0.1104}} & \textbf{\mygreen{-0.0019}} & \textbf{\mygreen{-0.014}} & \textbf{\myorange{+2.72\%}} & \textbf{\mygreen{-0.11\%}} & \textbf{\myorange{+3.46\%}} & \textbf{\mygreen{-1.50\%}} \\
        \midrule
        \multirow{3}{*}{ESFJ} & Original & 0.3491 & \cellcolor{gray!20} 0.3395 & 0.042 &  79.19\% & 44.35\% & 23.63\% & 70.91\% \\
         & Intervened & 0.5160 & \cellcolor{gray!20} 0.4785 & 0.012 & 79.86\% & 45.27\% & 26.67\% & 74.68\% \\
         & \textbf{Diff $\Delta$} & \textbf{\myorange{+0.1669}} & \cellcolor{gray!20} \textbf{\myorange{+0.1390}} & \textbf{\mygreen{-0.030}} & \textbf{\myorange{+0.67\%}} & \textbf{\myorange{+0.92\%}} & \textbf{\myorange{+3.04\%}} & \textbf{\myorange{+3.77\%}} \\
        \midrule
        \multirow{3}{*}{ISTP} & Original & 0.5126 & 0.7153 & \cellcolor{gray!20} 0.078 & 50.33\% & 47.42\% & 52.79\% & 41.95\% \\
         & Intervened & 0.4994 & 0.7080 & \cellcolor{gray!20} 0.042 & 51.10\% & 47.19\% & 52.50\% & 45.14\% \\
         & \textbf{Diff $\Delta$} & \textbf{\mygreen{-0.0132}} & \textbf{\mygreen{-0.0073}} & \cellcolor{gray!20} \textbf{\mygreen{-0.036}} & \textbf{\myorange{+0.77\%}} & \textbf{\mygreen{-0.23\%}} & \textbf{\mygreen{-0.29\%}} & \textbf{\myorange{+3.19\%}} \\
        \bottomrule
    \end{tabular}
}
\vspace{-15pt}
\end{table}

\section{Related work}
\label{sec:related_work}

\textbf{Myers-Briggs Type Indicator.}
Personality is a fundamental concept in psychology, referring to the dynamic integration of the totality of a person's subjective experience and behavior patterns \cite{kernberg2016personality}.
Various theories and models have been proposed to conceptualize and measure personality traits \cite{john1991big,eysenck1975manual,briggs1976myers}.
Myers-Briggs Type Indicator \cite{briggs1976myers} are 
based on Carl Jung's theory of psychological types. 
Two notable variants of the MBTI have been developed to meet specific research needs, \emph{i.e.}, MBTI-G \cite{carlson1985recent} and MBTI-M \cite{myers2003mbti}. 
These adaptations of the MBTI have been applied in various research to investigate the relationships between personality and other variables \cite{hough2005empirical,brown2009myers,higgs2001there,he2024afspp}.

\textbf{LLMs Personality Traits.}
Research suggests that LLMs exhibit unique personality traits that both resemble and differ from human personalities \cite{song2023have,lu2023illuminating,dorner2023personality,ai2024cognition}.
MBTI has been used to assess LLM personality \cite{pan2023llms,cui2023machine,ai2024cognition,song2024identifying}.
Specifically, there are two primary methods for assessing LLM personality: one is the direct application of human psychological scales to LLMs \cite{huang2023chatgpt,dorner2023personality,jiang2024evaluating}; the other is inferring personality traits based on language content generated by LLMs through specialized models \cite{song2024identifying,lu2023illuminating,wang2024incharacter}.
In terms of editing and shaping model personalities, researchers have proposed various methods to change LLMs personalities to suit different application scenarios and user needs, including prompt \cite{jiang2024evaluating,tan2024phantom}, role-playing \cite{wang2024incharacter,shao2023character}, model edit \cite{mao2023editing} and fine-tuning\cite{cui2023machine}.

\textbf{LLMs Alignment and Safety.}
The foundation of understanding Safety LLMs is established through existing research on AI governance~\cite{NIST, doi/10.2759/346720, AI_Act} and trustworthy AI~\cite{Liu2023trustworthy_ai, AI_Verify}. These studies provide guidance for identifying the core dimensions of trustworthiness in LLMs \cite{liu2023trustworthy, wang2024decodingtrust, sun2024trustllm, qian2024towards}. To this end, ensuring the alignment of LLMs with human values is crucial to mitigate and avoid potential societal safety risks. Many approaches have been proposed, including optimizing LLMs from human preferences \cite{ziegler2019fine, ouyang2022training, bai2022training, rafailov2023direct, lee2023rlaif, yang2023shadow} and self-alignment \cite{mita2020self, reid2022learning, madaan2023self}, This enables the LLM to identify and rectify the harmfulness of its outputs, thereby fostering greater alignment with societal values.

\section{Conclusion}
\label{sec:conclusion}

In this study, we discover that safety alignment can generally change LLMs' personality traits, and LLMs with different personality traits are differentially susceptible to jailbreaks. %
Meanwhile, we discover that LLMs' personality traits are closely related to their performance in safety capabilities such as toxicity, privacy, and fairness. 
Based on these findings, we experimentally demonstrate that editing LLMs' personality traits can enhance their safety performance, providing new insights for the development of LLM safety.
This study pioneers the exploration of LLM safety from a personality perspective. However, due to the complex correlation rather than causation between personality and safety in psychology \cite{lee2012correlation,careau2012performance}, there is still a need to further explore the more intrinsic relationship between personality traits and LLM safety.

\section*{Limitations}
\label{sec:limitations}

There are several limitations of this work. 
Firstly, our study focuses on 7B models due to the availability of both base and alignment models. Few 13B models offer both, limiting their representativeness. Thus, our main research centers on 7B models, with 13B models discussed in the appendix.
Secondly, we measure the MBTI traits of closed-source models without editing because there are no model weights to construct steering vectors, and the prompt methods are uncontrollable.
Thirdly, to reduce variables, we limited our study to three representative safety dimensions, ensuring a manageable scope and meaningful insights into the relationship between LLMs' personality traits and safety capabilities.

\section*{Broader Impact and Ethics Statement}
\label{sec:impact}

This study focuses on better understanding the relationship between personality traits and LLM safety. We emphasize that personality traits, assessed and edited in this study, do not imply any inherent value judgments. There are no ``good'' or ``bad'' personality traits, and our objective is only to enhance LLM safety. We strictly prohibit the intentional steering of models towards unsafe personality traits. All modifications are performed with the primary goal of improving model safety, ensuring that our work contributes positively to the development of ethical and trustworthy AI systems.

This research is carried out in a secure, controlled environment, ensuring the safety of real-world systems. Access to the most sensitive aspects of our experiments is limited to researchers with the proper authorization, who are committed to following rigorous ethical standards. 
These precautions are taken to maintain the integrity of our research and to mitigate any risks that could arise from the experiment's content.

\bibliographystyle{plain}
\bibliography{references}

\appendix

\newpage

\section{The Reliability of MBTI Assessment for LLMs}
\label{appendix:reliability}

\subsection{Setting Cases about Factors Affecting MBTI Assessment}

There are two types of the option order in the MBTI scale:
\begin{itemize}[leftmargin=*]
    \item \textbf{Exchange Option Description.} Following the settings of previous research \cite{wang2023large}, we exchange option descriptions while maintaining the order of label, \emph{i.e.}, changing \textit{A. Agree, B. Disagree} to \textit{A. Disagree, B. Agree}.
    \item \textbf{Exchange Option Label.} we exchange option labels while maintaining the order of descriptions, \emph{i.e}., changing \textit{A. Agree, B. Disagree} to \textit{B. Agree, A. Disagree}.
\end{itemize}

There are three factors that can affect MBTI assessment under each type of option order:
\begin{itemize}[leftmargin=*]
    \item \textbf{Option Label.} We set option labels in two forms: alphabets (\textit{e.g., A. Agree B. Disagree}) and numbers (\textit{e.g., 1. Agree 2. Disagree}), and examine their impact on the MBTI assessment results.
    \item \textbf{Instructions.} We provide two styles of instruction: (1) samples that answer contains option label and corresponding description (\textit{i.e., Question: Artificial intelligence cannot have emotions. A. Agree, B. Disagree. Your answer: B. Disagree}); (2) answer contains only option label without descriptions (\textit{i.e., Question: Artificial intelligence cannot have emotions. A. Agree, B. Disagree. Your answer: B}).
    \item \textbf{Language.} We use both Chinese and English versions of the MBTI-M questionnaire to assess the personality results of LLMs from different cultural backgrounds.
\end{itemize}

\subsection{Kappa coefficient of Option Order by Exchanging Option Labels}
\label{appendix:exchange_label}

Table~\ref{table:kappa-label} presents the kappa coefficients for various models under different settings in terms of the option order (\emph{i.e.}, the exchange option label). The results demonstrate that selecting the number as the option label consistently yields a higher kappa coefficient than the alphabet, suggesting that using ``number'' is the optimal selection for this factor. 
Additionally, the analysis reveals that the kappa coefficient on the MBTI results remains comparable regardless of whether the instructions include descriptions or not. This observation indicates that the option descriptions do not significantly impact the performance of the assessment in this setting. The kappa coefficient also exhibits consistency between the Chinese and English scales, which means that the MBTI results are not greatly affected by the assessment language.

\begin{table}[htbp]
\centering
\caption{Kappa coefficient of the option order (\underline{\textit{exchange option labels}}) in LLMs' MBTI assessment under three factors, respectively.}
\label{table:kappa-label}
\setlength{\tabcolsep}{1.5mm}
\scalebox{0.70}{
    \begin{tabular}{@{}cc|ccccccccccc@{}}
    \toprule
    \multicolumn{2}{c|}{\textbf{Factors}} & \textbf{Llama-2} & \textbf{Llama-3} & \textbf{Amber}  & \textbf{Gemma}  & \multicolumn{1}{l}{\textbf{Mistral}} & \multicolumn{1}{l}{\textbf{Baichuan}} & \textbf{Internlm} & \textbf{Internlm2} & \textbf{Qwen}   & \textbf{Qwen-1.5} & \textbf{Yi}     \\ \midrule
                                                                             & \textbf{number}    & \textbf{0.1017}  & \textbf{0.1846}  & \textbf{0.2733} & \textbf{0.0937} & 0.2493                               & \textbf{0.4003}                       & \textbf{0.1597}   & 0.2637             & 0.0203          & \textbf{0.2999}   & 0.0107          \\
    \multirow{-2}{*}{\begin{tabular}[c]{@{}c@{}}\textbf{Choice}\\ \textbf{Label}\end{tabular}} & \textbf{alphabet}  & 0.0143           & 0.094            & 0.0688          & 0.0092          & \textbf{0.335}                       & 0.2636                                & 0.058             & \textbf{0.5164}    & \textbf{0.3378} & 0.0827            & \textbf{0.6666} \\ \midrule
                                                                             & \textbf{w/ desc}   & \textbf{0.1285}  & \textbf{0.2184}  & \textbf{0.2352} & 0.0597          & \textbf{0.2678}                      & 0.319                                 & 0.1189            & 0.2274             & 0.0905          & 0.2725            & \textbf{0.0541} \\
    \multirow{-2}{*}{\textbf{ICL}}                                                    & \textbf{w/o desc}  & 0.0055           & 0.2163           & 0.024           & \textbf{0.1033} & 0.0139                               & \textbf{0.3491}                       & \textbf{0.2868}   & \textbf{0.2285}    & \textbf{0.2779} & \textbf{0.4045}   & 0.0553          \\ \midrule
                                                                             & \textbf{chinese}   & 0.0357           & 0.1535           & 0.2572          & \textbf{0.0453} & \textbf{0.2523}                      & 0.3924                                & 0.1006            & \textbf{0.2784}    & 0.0002          & \textbf{0.2704}   & 0.0137          \\
    \multirow{-2}{*}{\textbf{Language}}                                               & \textbf{english}   & \textbf{0.4702}  & \textbf{0.1635}  & \textbf{0.3983} & 0.041           & 0.1915                               & \textbf{0.4418}                       & \textbf{0.2622}   & 0.1299             & \textbf{0.2626} & 0.2353            & \textbf{0.4515} \\ \bottomrule
    \end{tabular}
}
\end{table}

\subsection{MBTI Mean and Standard Deviation with Number of Measurements}
\label{appendix:mbti_std}

\begin{figure}[htbp]
    \centering 
    \includegraphics[width=\textwidth]{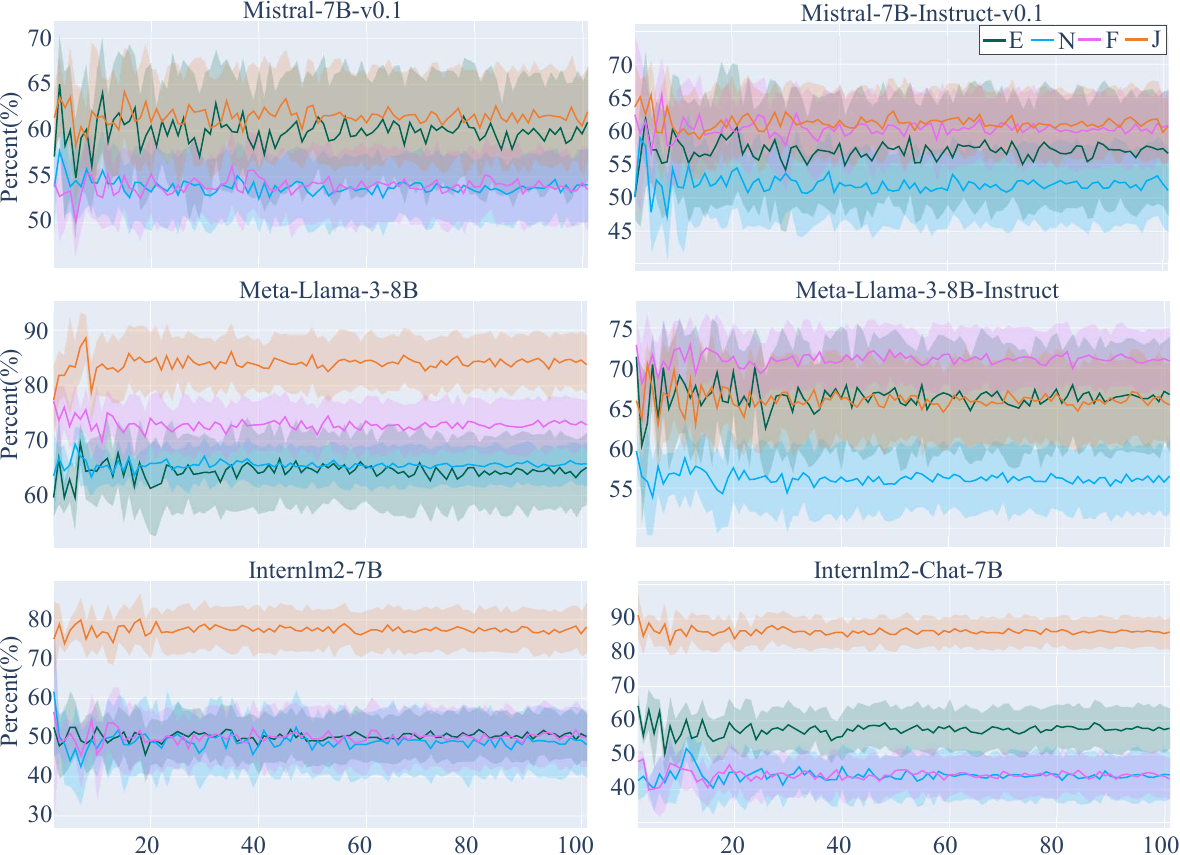}
    \caption{MBTI Results with Number of Assessments.}
    \label{fig:times}
\end{figure}

The results of personality assessments using psychological scales can vary between multiple-time measurements, even when administered to human subjects, and may change over time. As shown in Figure~\ref{fig:times}, there is always a certain standard deviation in the results of repeated MBTI assessments. Although the model's output tends to stabilize to some extent as the number of measurements increases, the standard deviation persists and never completely diminishes. This observation indicates that while employing multiple measurements can contribute to obtaining MBTI results, the influence of the standard deviation on the outcomes remains a notable factor.

The persistence of the standard deviation across multiple measurements highlights the inherent complexity and potential instability in capturing personality traits through psychological scales. 
To investigate this, We randomly shuffle the options in the scale before each assessment, conducting between 1 to 100 assessments for each model, and calculate the Kappa coefficient for each instance, as shown in Section~\ref{subsec:reliability} of the main text, thus verifying the reliability of the MBTI results. This method allows us to obtain reliable personality assessment outcomes in the presence of variability introduced by the option order.

\section{Boxplot of 30 times MBTI Assessment for More LLMs}
\label{appendix:morebox}

\subsection{Boxplots of Mindset}
\label{appendix:mindsetbox}

\begin{figure}[htbp]
    \centering 
    \includegraphics[width=\textwidth]{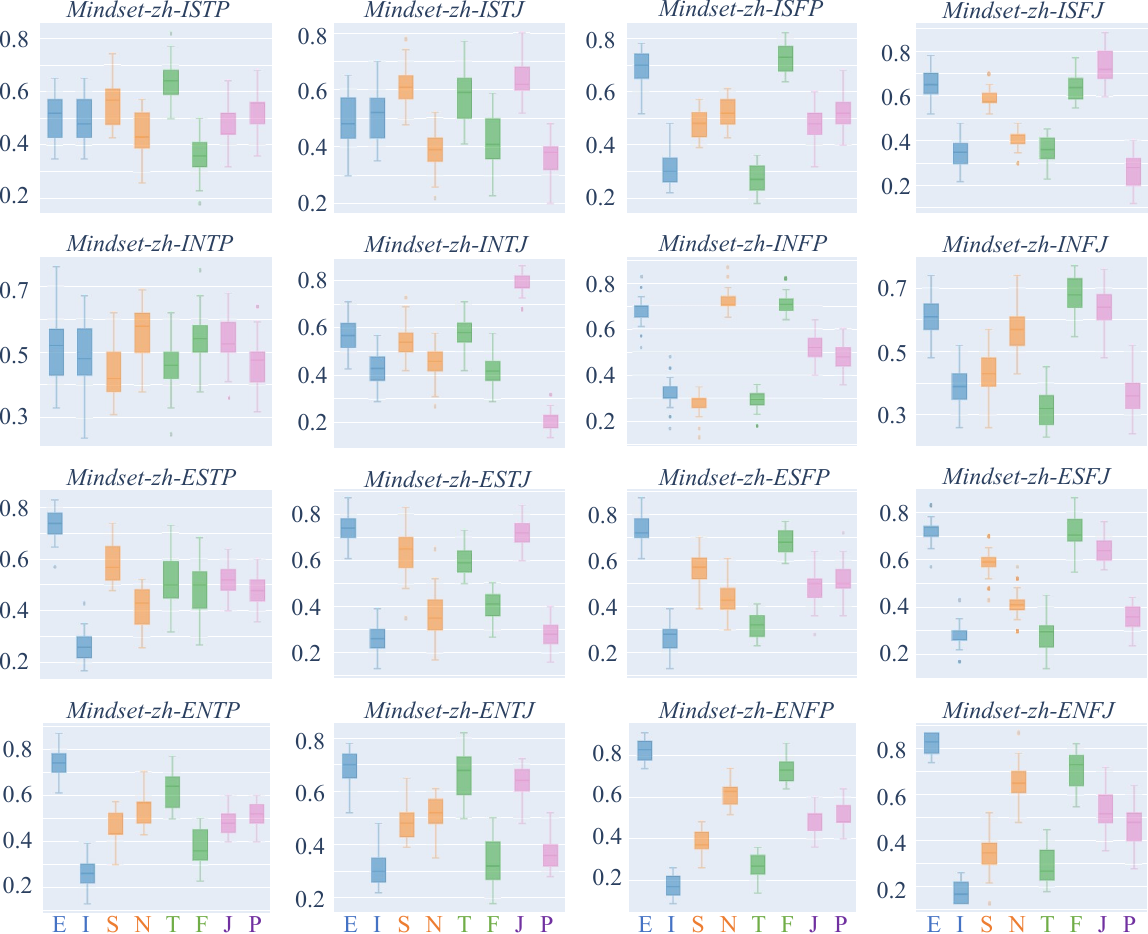}
    \caption{Boxplots of Mindset-zh.}
    \label{fig:zh-box}
\end{figure}

\begin{figure}[htbp]
    \centering 
    \includegraphics[width=\textwidth]{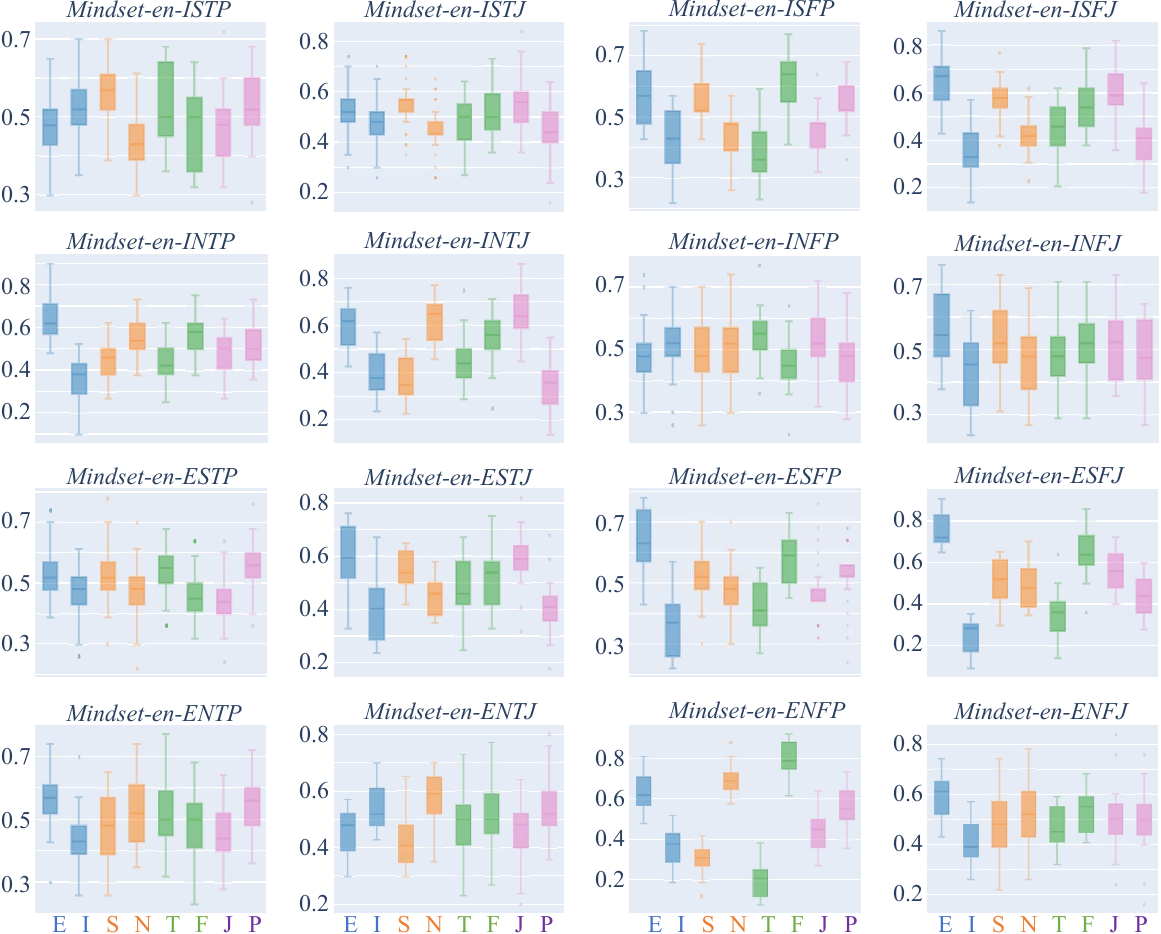}
    \caption{Boxplots of Mindset-en.}
    \label{fig:en-box}
\end{figure}

Based on the reliable MBTI assessment method described in Section~\ref{sec:personality}, we re-evaluate the personality of 32 LLMs provided by Mindset \cite{cui2023machine}. The assessment results for the 16 MBTI personality models in Mindset-zh (Chinese) and Mindset-en (English) are presented in Figure~\ref{fig:zh-box} and Figure~\ref{fig:en-box}, respectively. 
Our results demonstrate a significant alignment with the expected MBTI obtained through fine-tuning, indicating that our assessment method possesses significant discriminative power. In most cases, opposite personality pairs are clearly distinguishable, with scores distinctly located on one side of the 50\% threshold.

However, some discrepancies are observed, particularly in the Extraversion-Introversion (E-I) dimension, where models rarely exhibit introverted traits. This observation suggests that additional methods are needed to make the models more introverted. For certain models, specific personality traits, such as Sensing-Intuition (S-N), are not easily differentiated, with scores hovering around the 50\% mark. This finding implies that further refinement and shaping of these models' personalities may be required to achieve more distinct and well-defined traits.

\subsection{Boxplots of Larger LLMs}
\begin{figure}[htbp]
    \centering 
    \includegraphics[width=\textwidth]{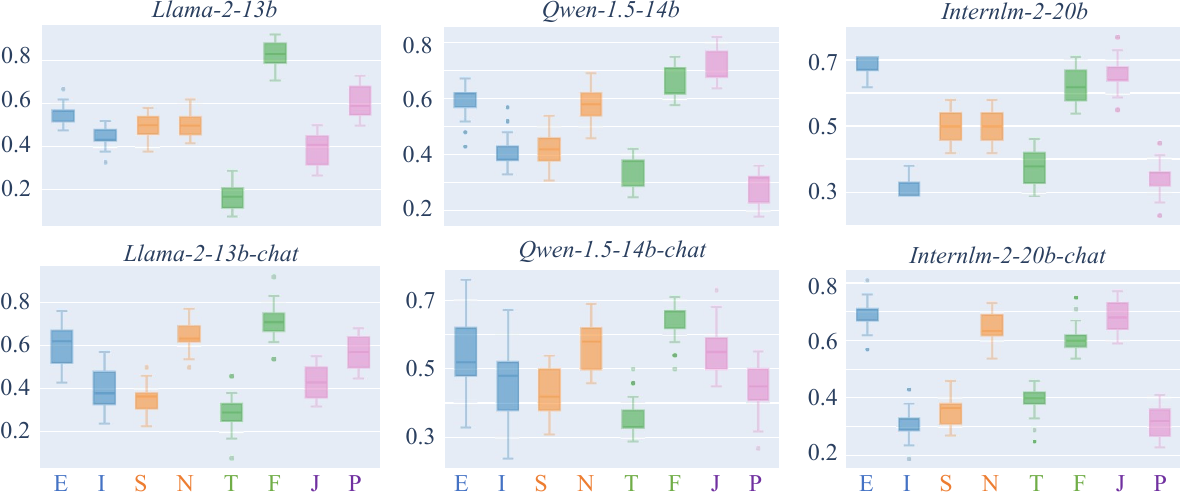}
    \caption{Boxplots of Llama-2-13b, Qwen-1.5-14b, and Internlm-2-20b.}
    \label{fig:larger-box}
\end{figure}

To further investigate the reliability and scalability of our MBTI-based personality analysis approach, we conduct 30 assessments on models with larger parameter scales, \emph{i.e.}, Llama-2-13b, Qwen-1.5-14b, and Internlm-2-20b. For each model, we get their personality traits using the MBTI scale and plot the results using box plots, as shown in Figure~\ref{fig:larger-box}. The box plots reveal that the MBTI personality dimensions of nearly all the tested models are significantly distinguishable, indicating that the models exhibit distinct personality profiles. For instance, in the case of Llama-2-13b, there is a substantial difference between the scores for the Feeling and Thinking (F-T) dimensions.

\section{Cultural Differences in the Context of Languages}
\label{appendix:culture}
In the realm of sociology, previous research~\cite{lee2007relations,ozanska2012has,veltkamp2013personality,chen2014does,church2016personality,allik2019culture} collectively suggests that language and culture significantly impact individual personality and behavior. These studies reveal that language is not merely a tool for communication but a crucial medium for shaping and expressing cultural identity, emotions, and social conduct. Furthermore, individuals may exhibit varying personality traits across different linguistic environments.

Thus, the observed differences in our experiments might be a reflection of these cultural and linguistic imprints on LLMs' learning process. In the context of LLMs, these findings suggest that the linguistic and cultural nuances embedded within a model's training data may shape its personality expressions and security behaviors. Moreover, the models' ability to adapt to security threats may be affected by emotional intelligence factors such as empathy and social awareness.

\section{Experiment Setting Details}
\label{appendix:setting}

\subsection{Mindset Model Selection in Four Dimensions of MBTI}
When analyzing the relationship between each of the four MBTI dimensions (E-I, N-S, T-F, J-P) and the three safety aspects (toxicity, privacy, and fairness) separately, we select models with significant differences in that personality dimension for analysis for each MBTI dimension. This selection process is based on the reliable MBTI results of our assessed mindset (see Appendix~\ref{appendix:mindsetbox}). The selection criteria are primarily twofold:

First, for each MBTI dimension (\emph{e.g.}, E-I), we select models that exhibit significant differences in that dimension for analysis. Specifically, we choose models with scores at the opposite ends of the dimension, i.e., those that clearly demonstrate either E or I, while avoiding models with scores in the middle. This ensures that the selected models have a clear distinction in that personality dimension.

Second, we also need to ensure that the number of models for each personality pair (\emph{e.g.}, E and I) is roughly balanced. This helps to balance the data, making the analysis results more reliable and statistically meaningful. If the number of models for a particular personality dimension is highly skewed, it may affect the reliability of the results.

\subsection{Controllable Editing with Steering Vector Technique}
In Section~\ref{subsec:change_mbti}, we conduct experiments on controllably editing the LLMs' personality traits (Figure~\ref{fig:steering_mbti}) based on Mindset-zh-ISTJ, Mindset-zh-ESFJ, and Mindset-en-ENFP. Additionally, we conduct experiments on changing the LLMs' safety capabilities (Table~\ref{tab:steeing_safe}) by changing the fairness of Mindset-zh-INFJ, the privacy of Mindset-zh-ESFJ, and the toxicity of Mindset-zh-ISTP. Notably, as observed in previous literature~\cite{privacy_fairness_1, privacy_fairness_2, qian2024towards}, there are trade-offs between different safety dimensions of a model (\textit{e.g.}, privacy-fairness trade-off~\cite{privacy_fairness_1}), it is challenging to observe a ``targeted'' change in a particular safety capability. Nevertheless, this does not undermine the conclusion that changing an LLM's safety capabilities impacts its personality traits.

When constructing steering vectors for safety datasets, we follow~\cite{li2023inferencetime, qian2024towards} to divide datasets into a development set and a test set in a $1:1$ ratio. The development set is used for constructing the steering vector, while the test set is used for evaluating the model's safety capabilities.

Regarding the Perplexity constraints mentioned in Section~\ref{subsec:steering_vector}, we follow the approach in~\cite{radford2019language} to calculate Perplexity on the LAMBADA~\cite{paperno-etal-2016-lambada} dataset. Following~\cite{qian2024towards}, we select a Perplexity threshold of $6$, considering intervention effects below this threshold as reasonable.

\section{Changes in MBTI after Safety Alignment for More LLMs}

\subsection{Llama-2 Series LLMs: Llama-2, Vicuna-1.5, and Tulu-2-dpo}
\label{appendix:llama-2}

Despite the overall trends, some models demonstrate personality shifts in the opposite direction, indicating potential interactions between alignment methods and the models' inherent characteristics. Our analysis of other Llama-2 aligned models (i.e., vicuna-1.5 and tulu-2-dpo) reveals that they also exhibit opposite personality shift patterns similar to Llama-2-chat, confirming the inherent model characteristics may cause individual models to deviate from the overall trends in personality changes.

\begin{figure}[htbp]
    \centering 
    \includegraphics[width=\textwidth]{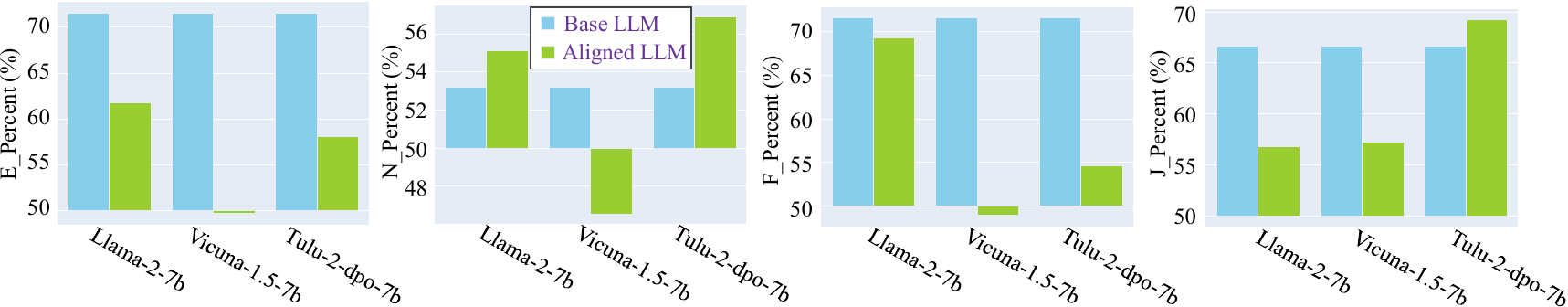}
    \caption{MBTI of base and aligned LLMs (Llama-2-7b, Vicuna-1.5-7b, and Tulu-2-dpo-7b).}
    \label{fig:llama}
\end{figure}

\subsection{Larger LLMs: Llama-2-13b, Qwen-1.5-14b, and Internlm-2-20b}
\label{appendix:largealign}

\begin{figure}[htbp]
    \centering 
    \includegraphics[width=\textwidth]{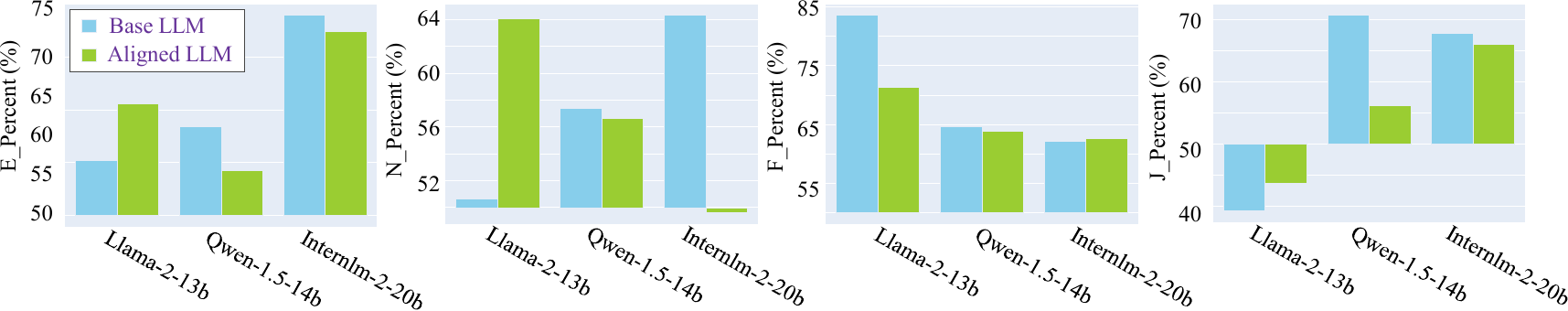}
    \caption{MBTI of base and aligned LLMs (Llama-2-13b, Qwen-1.5-14b, and Internlm-2-20b).}
    \label{fig:larger-align}
\end{figure}

We further assess larger LLMs, including Llama-2-13b, Qwen-1.5-14b, and Internlm-2-20b, analyzing their MBTI changes before and after alignment. 
The results, presented in Figure~\ref{fig:larger-align}, showcase the changes in personality dimensions observed in each model. However, due to the limited number of available models from the community at the corresponding parameter scales (13B, 14B, 20B), conducting a comprehensive statistical analysis of these findings remains challenging. To further advance research on LLM safety from the personality perspective, we strongly encourage increased open-sourcing efforts from the AI community. Researchers can explore the implications of personality traits on the safe development and deployment of LLMs.

\section{Experiments Compute Resources}
\label{appendix:resource}

All experiments in this study were conducted using NVIDIA A100 GPUs with 80GB memory. We perform MBTI personality assessment on the following models: 32 Mindset models, 22 base and align models with 7B parameters (11 pairs), 6 models with larger parameter sizes, and ChatGPT model, totaling 61 models. Storing a model with 7B parameters typically requires approximately 14GB of memory. Performing a single MBTI assessment on one model takes about 1 minute, with an estimated 100 hours for a complete single assessment process.

\end{document}